\documentclass{article} 
\usepackage{iclr2027_conference,times}

\usepackage{amsmath,amsfonts,bm}

\def\eqref#1{equation~\ref{#1}}

\def\1{\bm{1}}

\DeclareMathAlphabet{\mathsfit}{\encodingdefault}{\sfdefault}{m}{sl}
\SetMathAlphabet{\mathsfit}{bold}{\encodingdefault}{\sfdefault}{bx}{n}

\usepackage{siunitx}
\usepackage{etoolbox} 
\robustify\bfseries   
\usepackage{hyperref}
\usepackage{url}
\usepackage{booktabs}
\usepackage{graphicx} 
\usepackage{float} 
\usepackage{subfigure} 
\usepackage{makecell}
\usepackage{multirow} 
\usepackage[table]{xcolor}
\usepackage{array}
\usepackage{wrapfig}

\title{WeatherBridge: Forecast-Aware Multi-Source Input Adaptation for Pretrained Weather Forecasting Models}

\author{Mutian Xi$^{1,2}$\thanks{These authors contributed equally to this work.}, Xinyu Su$^{1,2}$\footnotemark[1], Xiaohui Zhong$^{1,2,3}$\thanks{Corresponding author. \texttt{zhong\_xiaohui@fudan.edu.cn, lihao\_lh@fudan.edu.cn}} \&  Hao Li$^{1,2,3}$\footnotemark[2] 
\\
$^{1}$Artificial Intelligence Innovation and Incubation Institute, Fudan University\\
$^{2}$Shanghai Academy of AI for Science\\
$^{3}$FuXi Intelligent Computing Technology Co., Ltd.\\
}

\newcommand{\model}{WeatherBridge}

\iclrfinalcopy 
\begin{document}

\maketitle

\begin{abstract}
Data-driven weather forecasting models are commonly trained on ERA5, yet operational deployment must use timely initial states from physics-based numerical weather prediction (NWP) systems whose distributions differ substantially from ERA5.
Direct initialization with these states can degrade autoregressive forecasts, motivating input adaptation that keeps the pretrained forecasting model unchanged.
Here, we introduce WeatherBridge, a unified, source-conditioned adapter that maps heterogeneous operational NWP initial states to ERA5 reference states without updating the pretrained forecasting model.
WeatherBridge predicts the structured residual corrections with a Swin-UNet and injects source embeddings across multiple spatial scales, allowing the model to capture shared atmospheric structure and source-specific discrepancies.
Its training objective combines latitude-weighted ERA5 reconstruction with One-step Rollout Alignment (ORA), which backpropagates one-step forecast error through a frozen FuXi model to optimize the adapted states for both initial state fidelity and downstream forecast compatibility.
We evaluate WeatherBridge using daily GFS, HRES, and CMA initializations from 1 January to 22 November 2025 across six variables and lead times up to 240 hours.
Relative to raw source initialization of FuXi, WeatherBridge reduces the mean RMSE of FuXi over the 6- 240-hour rollout for 17 of 18 source-variable combinations, with a median reduction of 10.2\% and a maximum reduction of 46.3\%.
The adapter trained with frozen FuXi also improves FengWu forecasts on GFS inputs without retraining either the adapter or FengWu. These results demonstrate that forecast-aware input adaptation can improve forecasts from heterogeneous operational NWP inputs and transfer across pretrained ERA5-based forecasting models.
\end{abstract}

\section{Introduction}

Data-driven weather forecasting models have emerged as promising alternatives to traditional numerical weather prediction (NWP) models for global medium-range weather forecasting.
State-of-the-art models learn atmospheric evolution from historical data and can match or outperform conventional NWP models at substantially lower inference cost \citep{lam2023graphcast,bi2023pangu,chen2023fengwu,chen2023fuxi,lam2023graphcast,bodnar2025aurora}.
Most of these models are trained on ERA5, whose data quality, global coverage, temporal consistency, public availability, and comprehensive set of multilevel variables have made it a standard benchmark for data-driven weather forecasting \citep{hersbach2020era5}.
However, operational forecasting requires timely initial conditions from NWP systems such as GFS, HRES, and CMA because ERA5 is available only after a delay of approximately five days.
Differences in data assimilation and numerical modeling cause these operational fields to exhibit statistical and structural characteristics that differ substantially from those of ERA5.
Consequently, directly initializing an ERA5-trained model with operational NWP fields can significantly degrade forecast performance.

\begin{figure}[t]
\centering
\includegraphics[width=0.95\linewidth]{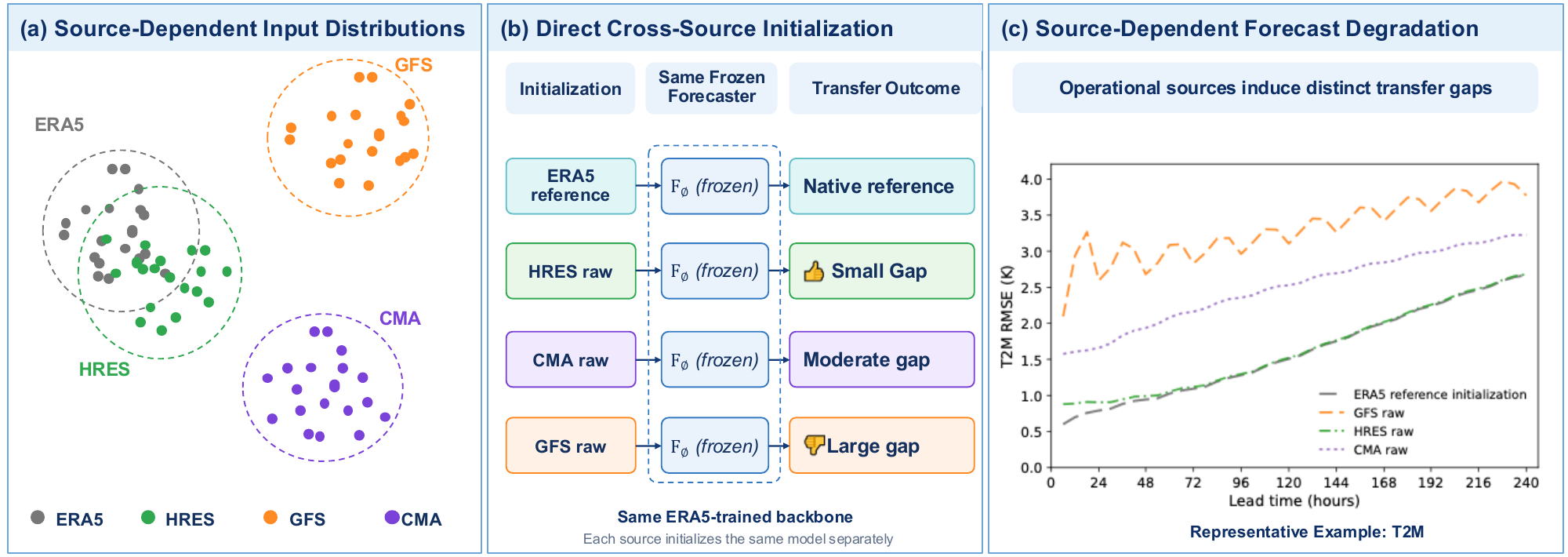}
\caption{Effect of operational-source initialization on a frozen
ERA5-trained forecaster.
(a) Schematic illustration of source-dependent input
distribution differences.
(b) Initialization of the same frozen forecaster with ERA5
or raw operational fields, without adaptation.
(c) T2M RMSE of FuXi initialized with ERA5, GFS, HRES, and CMA,
with all forecasts verified against ERA5.
The forecast errors vary across initialization sources.}
\label{fig:motivation}
\end{figure}

This input distribution shift is especially consequential for autoregressive forecasting.
Models such as FuXi, GraphCast, and Pangu-Weather implicitly rely on the variable definitions, normalization statistics, spatial structures, and multivariate relationships in ERA5.
Initial fields from GFS, HRES, and CMA exhibit source-specific discrepancies in these characteristics.
When such fields are fed directly to a frozen ERA5-trained model, the model operates outside its training distribution, and differences in the initial state can propagate and amplify through the autoregressive rollout.
Figure~\ref{fig:motivation} illustrates this effect for FuXi.
Initialization with raw GFS or CMA yields higher RMSE than ERA5 initialization at all lead times on variable 2 meter temperature (t2m).
HRES initialization begins closer to the ERA5 reference, and the performance gap generally narrows at longer lead times as forecast errors saturate.
Therefore, the extent of the degradation depends strongly on the discrepancy between each operational NWP source and ERA5.

Fine-tuning is one approach to adapting pretrained forecasting
models to different initialization data. For example, FuXi-GRAPES fine-tunes FuXi on CMA reanalysis
data~\citep{zhang2025fuxi}, while operational GraphCast is fine-tuned on ECMWF HRES analysis~\citep{deepmind2024graphcast}. Changes to operational systems can also prompt further
training, as illustrated by AIFS v2 following the IFS Cycle 50r1 upgrade~\citep{ecmwf2026ifs50r1,ecmwf2026aifsv2}.
We explore a complementary approach that trains a shared input adapter while keeping the pretrained forecasting
model unchanged.

We introduce WeatherBridge, a unified, source-conditioned adapter that maps heterogeneous operational NWP fields to ERA5 reference states while leaving the pretrained weather forecasting models unchanged. 
WeatherBridge represents the NWP-to-ERA5 mapping as a structured residual correction learned by a shared Swin-UNet.
To account for systematic differences among GFS, HRES, and CMA, WeatherBridge injects a learnable source embedding into the network at multiple spatial scales.
This hierarchical conditioning enables a single model to capture shared atmospheric structure while retaining NWP-specific corrections. 

Crucially, matching ERA5 at the initial time does not by itself guarantee that an adapted state will support accurate autoregressive forecasts.
Therefore, WeatherBridge combines latitude-weighted ERA5 reconstruction with One-step Rollout Alignment (ORA).
ORA propagates one-step forecast error through a frozen forecasting model, encouraging the adapter to produce states that are compatible with the model's learned dynamics and close to the corresponding ERA5 fields.
In this study, we train WeatherBridge using the frozen FuXi model and evaluate the resulting adapter on daily GFS, HRES, and CMA initializations over lead times of up to 240 hours.
Relative to direct initialization with raw source fields, WeatherBridge reduces FuXi's mean rollout RMSE for 17 of the 18 source-variable combinations, with reductions of up to 46.3\%.
When applied unchanged to the independently trained FengWu model, WeatherBridge also reduces RMSE and increases ACC for all six evaluated GFS variables at both 6 and 240 hours.

\textbf{Our contributions are as follows:}
\begin{itemize}
    \item We propose WeatherBridge, a unified, source-conditioned adapter that maps heterogeneous operational NWP initial fields to ERA5 reference states, allowing a frozen ERA5-trained forecasting model to use non-ERA5 inputs without fine-tuning.
    \item We introduce a hierarchical source conditioning mechanism that injects learned source embeddings at multiple spatial scales, allowing a single shared adapter to account for source-dependent discrepancies when mapping operational NWP inputs to ERA5 reference states.
    \item We introduce a dual-target training objective that combines latitude-weighted ERA5 reconstruction with One-step Rollout Alignment (ORA) through the frozen forecasting model, jointly promoting initial state fidelity and downstream forecast performance.
    \item We evaluate WeatherBridge on three operational NWP sources (GFS, HRES, CMA), six variables, and forecast lead times up to 240 h. The adapter improves FuXi forecasts and transfers unchanged to FengWu (another ERA5-trained forecasting model).
\end{itemize}

\section{Related Work}
\label{gen_inst}

\paragraph{ERA5-Trained AI-based Weather Forecasting Models.}

Recent AI-based systems, such as GraphCast, Pangu-Weather, FuXi, FourCastNet, FengWu, GenCast, and AIFS, have shown competitive medium-range forecast skill by learning global atmospheric evolution from reanalysis or operational analysis data \citep{lam2023graphcast,bi2023pangu,chen2023fuxi,pathak2022fourcastnet,chen2023fengwu,price2024gencast,lang2024aifs}. 
Despite their architectural differences, these models generally perform most reliably when their initial states follow the distribution encountered during training.
Therefore, operational deployment with a different NWP source may require source-specific adaptation.
WeatherBridge addresses the distribution mismatch at the input interface by mapping heterogeneous operational initial states to the ERA5 reference domain while leaving the forecasting model unchanged.

\paragraph{Bias Correction and Statistical Post-Processing.}

An alternative to model fine-tuning is to correct systematic errors in in NWP or AI weather forecasts using bias correction, model output statistics (MOS), quantile mapping, and AI-based post-processing.
These methods typically target particular variables, locations, or products \citep{vannitsem2021statistical,huang2024cmip6}. 
WeatherBridge transforms the initial state supplied to a pretrained forecasting model, with the objective of improving both its agreement with ERA5 and the resulting autoregressive forecasts.
To this end, its training objective propagates forecast error through the frozen FuXi model, explicitly accounting for how the adapted state affects downstream forecast performance.

\paragraph{Earth-System Foundation Models and Heterogeneous Data.}

Earth-system foundation models such as Aurora and ClimaX accommodate multiple data sources, resolutions, variable sets, and prediction tasks through large-scale pretraining and task-specific adaptation \citep{bodnar2025aurora,nguyen2023climax}.
This strategy seek to make the model broadly transferable by exposing it to heterogeneous data during training.
WeatherBridge pursues a complementary pathway.
Rather than requiring a single forecasting model to accommodate multiple NWP data distributions, it preserves the ERA5-trained backbone and learns a source-conditioned mapping from the operational NWP initial states to ERA5 reference state.
The distinction is between building heterogeneity into the forecasting model and resolving heterogeneity before the model is used.

\paragraph{Model Reprogramming and Input-Side Adaptation.}
WeatherBridge is also related to model reprogramming, which redirect a frozen or lightly modified pretrained model to new tasks or domains through input transformations, prompts, masks, or lightweight modules \citep{ye2024reprogrammability,cai2024masks}. 
Both approaches keep the major pretrained model unchanged and adapt its interface to a new input distribution.
However, WeatherBridge operates on physically meaningful, multivariate, spatio-temporally continuous meteorological states rather than image patches or tokens.
In addition, WeatherBridge is trained not only to reconstruct a reference state but also to improve the subsequent autoregressive forecast.

\section{Methodology}

We present \model, a plug-and-play adapter that transforms heterogeneous operational NWP initial states into ERA5-aligned states for a frozen pretrained weather forecasting model. In this work, we instantiate the forecasting backbone with FuXi~\citep{chen2023fuxi} while keeping its architecture and parameters unchanged throughout adapter training. 
Figure~\ref{fig:framework} illustrates the overall framework. We first formalize the input adaptation problem in Section~\ref{subsec:problem_form}. We then describe the source-conditioned adapter architecture in Section~\ref{subsec:arch}, which builds on a residual Swin-UNet and injects source identity hierarchically at multiple feature resolutions. Finally, we introduce the forecast-aware training strategy in Section~\ref{subsec:training}, which combines latitude-weighted ERA5 reconstruction with One-step Rollout Alignment (ORA) while keeping weather forecasting backbone frozen.


\subsection{Problem Formulation}
\label{subsec:problem_form}
Let $\mathcal{S}=\{\mathrm{GFS},\mathrm{HRES},\mathrm{CMA}\}$ denote the set of operational NWP sources. We index atmospheric states at 6-hour intervals, such that $t-1$, $t$, and $t+1$ denote three consecutive valid times. For each source $s\in\mathcal{S}$, let $x_t^s$ and $x_t^E$ denote the source and ERA5 states, respectively, at the same valid time $t$. Both are represented as tensors in $\mathbb{R}^{C\times H\times W}$, where $C=70$ is the number of atmospheric channels and $(H,W)=(721,1440)$ correspond to the latitude and longitude dimensions of the global $0.25^{\circ}$.

Given a single source state, its source identity, and the associated temporal information, \model\ learns the frame-wise mapping
{
\small
\begin{equation}
    \hat{x}_t^E = F_{\theta}^{\mathrm{ada}}(x_t^s,s,t),
    \label{eq:WeatherBridge_mapping}
\end{equation}
}
where $\hat{x}_t^E$ is the reference state produced from the source state at time $t$. The adapter always operates on one atmospheric state at a time.

Let $F_{\phi}$ denote a forecasting model pretrained on ERA5, whose parameters $\phi$ remain fixed. Given two consecutive source states, the goal is to learn a frame-wise adapter $F_{\theta}^{\mathrm{ada}}$ such that the states obtained by independently adapting the two frames enable $F_{\phi}$ to accurately predict the subsequent atmospheric evolution. Formally, the desired property is
{
\small
\begin{equation}
\tilde{x}_{t+k}^{E}=F_{\phi}^{[k]}\!\left(
    \left[
        F_{\theta}^{\mathrm{ada}}(x_{t-1}^{s},s,t-1),
        F_{\theta}^{\mathrm{ada}}(x_{t}^{s},s,t)
    \right]
\right)
\approx x_{t+k}^{E},
\qquad k=1,\ldots,K,
\label{eq:task_target}
\end{equation}
}
where $[\cdot,\cdot]$ denotes temporal concatenation, and $F_{\phi}^{[k]}$ denotes the atmospheric state predicted for valid time $t+k$ through a $k$-step autoregressive rollout of the frozen forecasting model.

\subsection{Source-Conditioned Adapter Architecture}
\label{subsec:arch}

\begin{figure}[t]
\centering
\includegraphics[width=\linewidth]{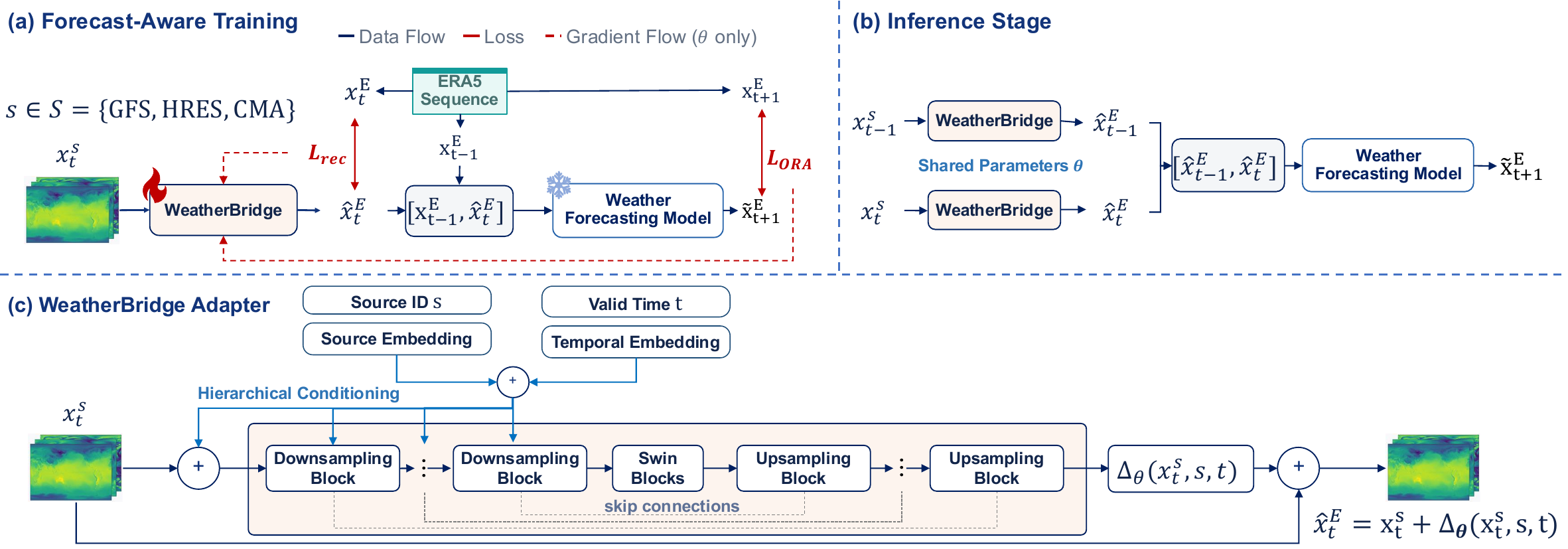}
\caption{{Overview of \model.}
\textit{(a) Forecast-aware training.}
Given a source state $x_t^s$, \model\ maps it to a reference-domain state $\hat{x}_t^E$. The adapted state and the preceding ERA5 state $x_{t-1}^E$ are fed into the frozen weather forecasting model to predict $\hat{x}_{t+1}^E$. The adapter is optimized using the reconstruction loss $\mathcal{L}_{\mathrm{rec}}$ at time $t$ and the one-step rollout loss $\mathcal{L}_{\mathrm{ORA}}$ at time $t+1$, while the forecasting model remains frozen.
\textit{(b) Inference stage.}
Two consecutive source states, $x_{t-1}^s$ and $x_t^s$, are mapped by \model\ with shared parameters. The resulting state pair is then passed to the frozen weather forecasting model to predict $\hat{x}_{t+1}^E$.
\textit{(c) \model\ adapter.}
The source identity and valid time are encoded as embeddings and hierarchically injected into multiple downsampling stages of a residual Swin-UNet. The network predicts a source-aware and time-aware correction $\Delta_\theta(x_t^s,s,t)$, yielding $\hat{x}_t^E=x_t^s+\Delta_\theta(x_t^s,s,t)$.}
\label{fig:framework}
\end{figure}

Figure~\ref{fig:framework} provides an overview of \model\ and its integration with the frozen pretrained forecasting model. \model\ employs a residual Swin-UNet to transform each source state into the reference state. A hierarchical conditioning mechanism injects source identity at multiple feature resolutions, enabling the
shared adapter to learn source-specific corrections across spatial scales.

\paragraph{Residual Swin-UNet Backbone.}

Our adapter employs a residual Swin-UNet backbone to transform each source state into the reference state. A convolutional patch embedding first projects the atmospheric variables into a latent
feature space. The encoder stacks convolutional residual blocks with
strided downsampling to progressively aggregate spatial information,
while a window-based self-attention block at the bottleneck models
broader spatial dependencies at a reduced resolution.
The decoder uses interpolation-based upsampling and convolutional
residual blocks to reconstruct the atmospheric state. At each decoder
stage, the upsampled representation is fused with the corresponding
encoder feature through a skip connection, retaining spatial
information that may be weakened during downsampling. A patch-recovery
head then maps the decoded representation back to the original
variable space and spatial grid. Rather than predicting the complete
target state directly, the network estimates a residual correction to
the normalized source field:
{
\small
\begin{equation}
\hat{x}_t^E
=
F_\theta^{\mathrm{ada}}(x_t^s,s,t)
=
x_t^s+\Delta_\theta(x_t^s,s,t),
\label{eq:residual}
\end{equation}
}
where $\Delta_\theta$ denotes the correction predicted by the
encoder--decoder backbone.
Both $x_t^s$ and
$\Delta_\theta(x_t^s,s,t)$ are represented in the same normalized
variable space, and conversion back to physical units is performed
outside the adapter when required. This residual parameterization
provides a direct identity path from the source state to the adapted
state and represents the source-to-ERA5 transformation as a correction
to the original field.

\paragraph{Hierarchical Conditioning.}

\model\ shares the same encoder--decoder parameters across all NWP
sources while conditioning the network on the identity of each source.
Specifically, the categorical source identifier
$s\in\mathcal{S}$ is used to retrieve a source embedding from a
learnable lookup table, i.e., $e_s = E_{\mathrm{src}}[s]$. Here, $E_{\mathrm{src}}
\in
\mathbb{R}^{|\mathcal{S}|\times d_s}$, and  $d_s$ denotes the embedding dimension.
This embedding table is
optimized jointly with the adapter, enabling a single shared backbone
to retain source-specific information.

In addition to the source identity, we derive a fixed sinusoidal
temporal embedding $e_t$ from the seasonal phase of the valid time,
computed using the day of year and hour of day. This embedding provides
temporal metadata associated with the current atmospheric state and
does not introduce an additional input frame.

After being projected to the input channel dimension, the source and
temporal embeddings are broadcast over the spatial grid and added to
the source field before patch embedding. To preserve the conditioning information throughout the encoder hierarchy, both embeddings are projected and injected at every downsampling level. Given a
downsampling feature $h_l$ at level $l$, the conditioned representation is
computed as
{
\small
\begin{equation}
\bar{h}_l
=
h_l
+
\mathcal{P}_l^s(e_s)
+
\mathcal{P}_l^t(e_t),
\label{eq:hierarchical_conditioning}
\end{equation}
}
where $\mathcal{P}_l^s$ and $\mathcal{P}_l^t$ denote stage-specific
projections that match the channel dimension and spatial resolution
of $h_l$. Conditioning is applied throughout the encoder but not in
the decoder, allowing the shared backbone to model source-dependent
corrections at multiple spatial scales.

\subsection{Forecast-aware Training Strategy}
\label{subsec:training}

We train \model\ to map heterogeneous NWP source fields to adapted initial
fields that support downstream forecasting. Toward this goal, we use the
standard reference-state reconstruction loss
$\mathcal{L}_{\mathrm{rec}}$ as the base training objective, encouraging
the adapted initial field to approach its paired reference state.
However, reconstruction alone does not explicitly account for how the
adapted field affects the downstream forecasting model. We therefore
augment the base objective with our proposed One-step Rollout Alignment
(ORA) loss $\mathcal{L}_{\mathrm{ORA}}$, which propagates the one-step
forecast error through the frozen forecasting model. The resulting
training objective is
\begin{equation}
\mathcal{L}_{\mathrm{total}}
=
\mathcal{L}_{\mathrm{rec}}
+
\lambda_f\mathcal{L}_{\mathrm{ORA}},
\label{eq:total_loss}
\end{equation}
where $\lambda_f>0$ controls the contribution of the forecast-aware
ORA term.

Specifically, the reconstruction term is defined as a
latitude-weighted $\ell_1$ loss between the adapted initial field and
its paired ERA5 reference state:
{
\small
\begin{equation}
\mathcal{L}_{\mathrm{rec}}
=
\frac{1}{C}
\sum_{c=1}^{C}
\frac{
\sum_i w_i
\left|
\hat{x}_{t,i}^{(c)}
-
x_{t,i}^{E,(c)}
\right|
}{
\sum_i w_i
},
\qquad
w_i=\cos(\phi_i),
\label{eq:rec_loss}
\end{equation}
}
where $C$ is the number of atmospheric channels, $i$ indexes spatial
grid points, and $\phi_i$ denotes the latitude of grid point $i$. The latitude weights correct for unequal grid-cell areas, with both
$\hat{x}_t$ and $x_t^E$ represented in normalized space.

To introduce forecast-aware supervision, ORA evaluates the adapted
initial field through the frozen forecasting model. Given the adapter
output $\hat{x}_t^E$, we combine it with the preceding ERA5 reference
state to construct the two-frame input
$[x_{t-1}^E,\hat{x}_t^E]$. The frozen model then produces the one-step
forecast $\tilde{x}_{t+1}$.
Using the ERA5 reference state as the preceding frame isolates the
effect of the current adapted field without introducing an additional
error from the previous input frame. The forecast error is
back-propagated through $F_{\phi}$ to update the adapter parameters
only. 

For each guided channel, we first compute a latitude-weighted RMSE
over the spatial grid and then normalize it by the channel-specific
reference RMSE of FuXi under native ERA5 inputs. The resulting
normalized errors are averaged equally across the guided channels:
{
\small
\begin{equation}
\mathcal{L}_{\mathrm{ORA}}
=
\frac{1}{|\mathcal{C}_{\mathrm{ORA}}|}
\sum_{c\in\mathcal{C}_{\mathrm{ORA}}}
\frac{
\operatorname{RMSE}_{w}^{c}
\left(
\tilde{x}_{t+1}^{E},
x_{t+1}^{E}
\right)
}{
\operatorname{RMSE}_{\mathrm{ref}}^{c}
}.
\label{eq:ora_loss}
\end{equation}
}
where $\operatorname{RMSE}_{w}^{c}$ is the latitude-weighted RMSE for
channel $c$, and $\operatorname{RMSE}_{\mathrm{ref}}^{c}$ is obtained
by forecasting from the native ERA5 input
$[x_{t-1}^E,x_t^E]$. We apply ORA to
{
\small
\begin{equation}
\mathcal{C}_{\mathrm{ORA}} =
\{
\texttt{z500}, \texttt{z850}, \texttt{u200},
\texttt{t2m}, \texttt{u10m}, \texttt{v10m},
\texttt{msl}, \texttt{tp}, \texttt{t300}
\}.
\end{equation}
}
which spans upper-, middle-, lower-, and near-surface atmospheric
conditions together with precipitation. The corresponding reference
errors are reported in Table~\ref{tab:ref_rmse} in the Appendix.

At inference, the reference states are neither available nor required.
\model\ is applied independently to two consecutive states
$x_{t-1}^s$ and $x_t^s$ from the same operational source, producing
$\hat{x}_{t-1}^E$ and $\hat{x}_t^E$, respectively. The two adapted
initial fields are then used as the FuXi input
$[\hat{x}_{t-1}^E,\hat{x}_t^E]$. Thus, unlike ORA training, where the
preceding frame is provided by the ERA5 reference sequence, both
frames used for inference are derived exclusively from operational
source data.

\section{Experiments}
\subsection{Experimental Setup}

\paragraph{Data.} 
We use ERA5 as the target domain and GFS, HRES, and CMA as heterogeneous source domains. Each source field is paired with the ERA5 field at the same valid time. All datasets are represented as FuXi-compatible 70-channel global fields~\cite{chen2023fuxi} on a $0.25^{\circ}$ grid with $721 \times 1440$ points, including 5 variables at 13 pressure levels and 5 surface variables.
All input fields are normalized using the ERA5 statistics $\mu_{\mathrm{ERA5}}$ and $\sigma_{\mathrm{ERA5}}$.
For all datasets, samples are initialized once per day, with consecutive forecast states spaced at 6-hour intervals. We use data from 2022-01-01 to 2024-06-30 for training, from 2024-07-01 to 2024-12-31 for validation, and from 2025-01-01 to 2025-11-22 for testing.

\paragraph{Implementation details.}
\model\ contains 138.44M trainable parameters and uses a dropout rate of 0.1. With the FuXi backbone frozen, we train \model\ in PyTorch on four A100 GPUs using FSDP, mixed precision, and AdamW with $\beta_1=0.9$, $\beta_2=0.999$, and a weight decay of $2\times10^{-5}$. We use a global batch size of 32, with 8 samples per GPU, clip the gradient norm at 5.0, linearly warm up the learning rate to $1\times10^{-4}$ over the first 5 epochs, and then apply cosine decay to $1\times10^{-7}$ over the remaining 85 epochs.

\paragraph{Evaluation.}
Each test initialization is rolled out for 40 six-hour steps, corresponding to a forecast horizon of 240 hours. We compare three initialization paths: ERA5 $\rightarrow$ FuXi as the reference, raw source $\rightarrow$ FuXi as the direct-inference baseline, and raw source $\rightarrow$ \model\ $\rightarrow$ FuXi as the proposed approach. We do not retrain or fine-tune FuXi separately for individual sources, as full source-specific retraining would require updating all 1,556.72M parameters of the backbone for each source and is outside the scope of this work. At inference, the proposed path uses only operational source fields. Following FuXi~\citep{chen2023fuxi}, all three initialization paths are evaluated against ERA5. We report results for \texttt{z500}, \texttt{t2m}, \texttt{t850}, \texttt{ws10}, \texttt{ws850}, and \texttt{msl}, where \texttt{ws10} and \texttt{ws850} are computed as the magnitudes of the horizontal wind vectors from the corresponding U and V components at 10\,m and 850\,hPa, respectively.
We evaluate \model\ at both the initialization and rollout forecasting levels. PSNR and SSIM provide supplementary measures of initial-state agreement with ERA5. Our primary evaluation uses latitude-weighted RMSE and ACC over the 240-hour rollout to assess downstream forecast skill. The definitions of all metrics are provided in Appendix~\ref{app:metrics}.We additionally report the ORA loss as a forecast-aware diagnostic for every configuration in the Appendix, and Table~\ref{tab:ref_rmse} lists the per-channel reference errors used to normalize it.

\subsection{Weather Forecasting Performance}
\label{subsec:weather_forecasting}

\begin{figure}[!t]
\centering
\includegraphics[width=\textwidth]{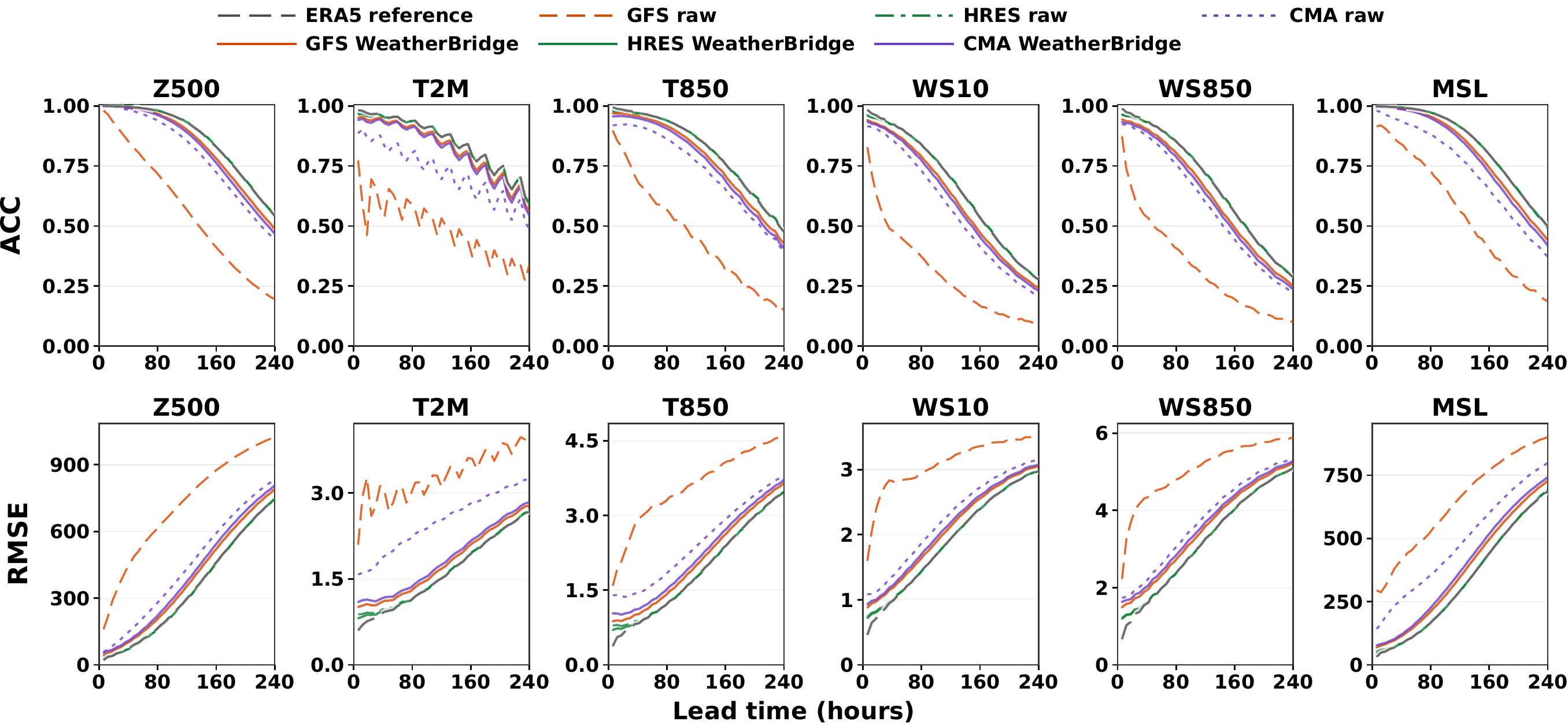}
\caption{Forecasting performance on FuXi over the 240-hour rollout for the six evaluation variables. Higher ACC and lower RMSE indicate better forecasting performance. Solid curves denote the \model-aligned initialization, while dashed curves of the same color denote the raw-source initialization for GFS (orange), HRES (green), and CMA (purple). The gray dashed curve denotes the ERA5 reference initialization.}
\label{fig:WeatherBridge_effect}
\end{figure}

\begin{table*}[t]
\caption{Per-source and per-variable results on FuXi. PSNR and SSIM assess initial-state alignment with ERA5, while mean RMSE is averaged over the 40-step rollout up to 240~h. ``$\uparrow$'' and ``$\downarrow$'' indicate that higher and lower values are better, respectively. $\Delta$ denotes the relative improvement of the \model-aligned result over the raw-source result; negative values indicate degradation.}
\label{tab:varwise}
\centering
\footnotesize
\setlength{\tabcolsep}{3pt}
\resizebox{\textwidth}{!}{%
\begin{tabular}{
@{}lc|
cc>{\columncolor{gray!15}}c|
cc>{\columncolor{gray!15}}c|
cc>{\columncolor{gray!15}}c@{}
}
\toprule
\multirow{2}{*}{Source}
& \multirow{2}{*}{Var.}
& \multicolumn{3}{c|}{PSNR $\uparrow$}
& \multicolumn{3}{c|}{SSIM $\uparrow$}
& \multicolumn{3}{c}{Mean RMSE $\downarrow$} \\
\cmidrule(lr){3-5}
\cmidrule(lr){6-8}
\cmidrule(lr){9-11}
& & raw & \model & $\Delta$
  & raw & \model & $\Delta$
  & raw & \model & $\Delta$ \\
\midrule

\multirow{6}{*}{GFS}
& Z500  & 44.78 & 46.68 & 4.24\%  & 0.9999 & 0.9999 & 0.00\% & 719.36 & 386.46 & 46.28\% \\
& T2M   & 31.45 & 34.27 & 8.97\%  & 0.9943 & 0.9971 & 0.28\% & 3.31   & 1.78   & 46.22\% \\
& T850  & 29.61 & 33.68 & 13.75\% & 0.9922 & 0.9967 & 0.45\% & 3.60   & 2.11   & 41.39\% \\
& WS10  & 21.89 & 23.77 & 8.59\%  & 0.9409 & 0.9609 & 2.13\% & 3.09   & 2.08   & 32.69\% \\
& WS850 & 23.50 & 24.18 & 2.89\%  & 0.9529 & 0.9596 & 0.70\% & 5.06   & 3.51   & 30.63\% \\
& MSL   & 23.95 & 35.34 & 47.56\% & 0.9442 & 0.9963 & 5.52\% & 642.47 & 372.90 & 41.96\% \\
\midrule

\multirow{6}{*}{HRES}
& Z500  & 50.03 & 50.02 & -0.02\% & 1.0000 & 1.0000 & 0.00\% & 338.17 & 338.33 & -0.05\% \\
& T2M   & 36.41 & 36.99 & 1.59\%  & 0.9982 & 0.9985 & 0.03\% & 1.64   & 1.63   & 0.61\% \\
& T850  & 36.17 & 36.81 & 1.77\%  & 0.9981 & 0.9984 & 0.03\% & 1.92   & 1.90   & 1.04\% \\
& WS10  & 24.94 & 25.43 & 1.96\%  & 0.9711 & 0.9736 & 0.26\% & 1.93   & 1.92   & 0.52\% \\
& WS850 & 25.72 & 25.98 & 1.01\%  & 0.9715 & 0.9734 & 0.20\% & 3.23   & 3.23   & 0.00\% \\
& MSL   & 39.09 & 39.60 & 1.30\%  & 0.9985 & 0.9987 & 0.02\% & 330.32 & 329.38 & 0.28\% \\
\midrule

\multirow{6}{*}{CMA}
& Z500  & 45.09 & 44.53 & -1.24\% & 0.9998 & 0.9998 & 0.00\% & 446.81 & 403.29 & 9.74\% \\
& T2M   & 30.87 & 33.61 & 8.88\%  & 0.9936 & 0.9966 & 0.30\% & 2.50   & 1.84   & 26.40\% \\
& T850  & 31.13 & 33.19 & 6.62\%  & 0.9943 & 0.9965 & 0.22\% & 2.46   & 2.21   & 10.16\% \\
& WS10  & 20.33 & 23.13 & 13.77\% & 0.9278 & 0.9544 & 2.87\% & 2.26   & 2.13   & 5.75\% \\
& WS850 & 23.19 & 23.54 & 1.51\%  & 0.9477 & 0.9528 & 0.54\% & 3.76   & 3.60   & 4.26\% \\
& MSL   & 29.29 & 35.09 & 19.80\% & 0.9855 & 0.9963 & 1.10\% & 486.70 & 388.67 & 20.14\% \\
\bottomrule
\end{tabular}
}
\end{table*}

Figure~\ref{fig:WeatherBridge_effect} compares the ACC and RMSE trajectories produced by the three initialization paths over the 240-hour rollout. The benefit of \model\ is most pronounced for GFS. Directly initializing FuXi with raw GFS fields leads to substantially lower ACC and higher RMSE across all six variables, with the gap generally increasing over the forecast horizon. After alignment, the GFS curves move consistently toward the ERA5 reference for both metrics, and the improvement persists throughout the rollout. CMA exhibits a similar but smaller improvement, with the clearest gains for T2M and MSL. In contrast, the raw and aligned HRES curves remain close to each other and to the ERA5 reference, indicating limited room for further improvement.

The mean rollout RMSE in Table~\ref{tab:varwise} quantifies this source-dependent pattern. \model\ reduces the mean RMSE for 17 of the 18 source-variable pairs. For GFS, the reductions range from 30.6\% to 46.3\% across the six variables. CMA also improves consistently, with reductions ranging from 4.3\% to 26.4\%, including the largest gains of 26.4\% for T2M and 20.1\% for MSL. The changes for HRES are considerably smaller: five variables show RMSE reductions of at most 1.04\%, while Z500 exhibits a slight increase of 0.05\%. Overall, the results show that \model\ provides the greatest forecasting benefit when direct initialization causes a substantial loss of forecast skill, while largely preserving the performance of sources that are already compatible with the ERA5-trained backbone.

\begin{table*}[t]
\centering
\caption{FengWu results on GFS at 240\,h with the FuXi-trained
adapter, neither retrained.
$\uparrow$/$\downarrow$: higher/lower is better.
$\Delta$: relative gain over raw-source initialization.}
\small
\label{tab:fengwu_transfer}
\setlength{\tabcolsep}{2.5pt}
\begin{tabular}{
@{}l|
rr>{\columncolor{gray!15}}r|
rr>{\columncolor{gray!15}}r@{}
}
\toprule
\multirow{2}{*}{Var.}
& \multicolumn{3}{c|}{RMSE $\downarrow$}
& \multicolumn{3}{c}{ACC $\uparrow$} \\
\cmidrule(lr){2-4}
\cmidrule(lr){5-7}
& Raw & Adapted & $\Delta$ (\%)
& Raw & Adapted & $\Delta$ (\%) \\
\midrule
Z500
& 1023.32 & 952.65 & 6.91
& 0.0732 & 0.1748 & 138.68 \\
T2M
& 3.93 & 3.23 & 17.77
& 0.3584 & 0.4365 & 21.78 \\
T850
& 4.81 & 4.08 & 15.25
& 0.0783 & 0.1717 & 119.27 \\
WS10
& 3.88 & 3.86 & 0.61
& 0.0479 & 0.0907 & 89.50 \\
WS850
& 6.29 & 5.91 & 6.08
& 0.0322 & 0.0873 & 171.17 \\
MSL
& 867.90 & 727.29 & 16.20
& 0.0952 & 0.2912 & 205.81 \\
\bottomrule
\end{tabular}
\end{table*}

\paragraph{Transfer to FengWu.}
The adapter trained with frozen FuXi is applied directly to FengWu
without retraining either the adapter or FengWu.
On GFS inputs, adaptation reduces the 240-hour RMSE by
0.61--17.77\% and increases ACC for all six evaluated variables
(Table~\ref{tab:fengwu_transfer}).
Results at 6\,h are provided in Appendix~\ref{app:subsec:fengwu}.

\subsection{Initial-State Alignment}
\label{subsec:state_align}
\begin{figure}[t]
\centering
\includegraphics[width=\textwidth]{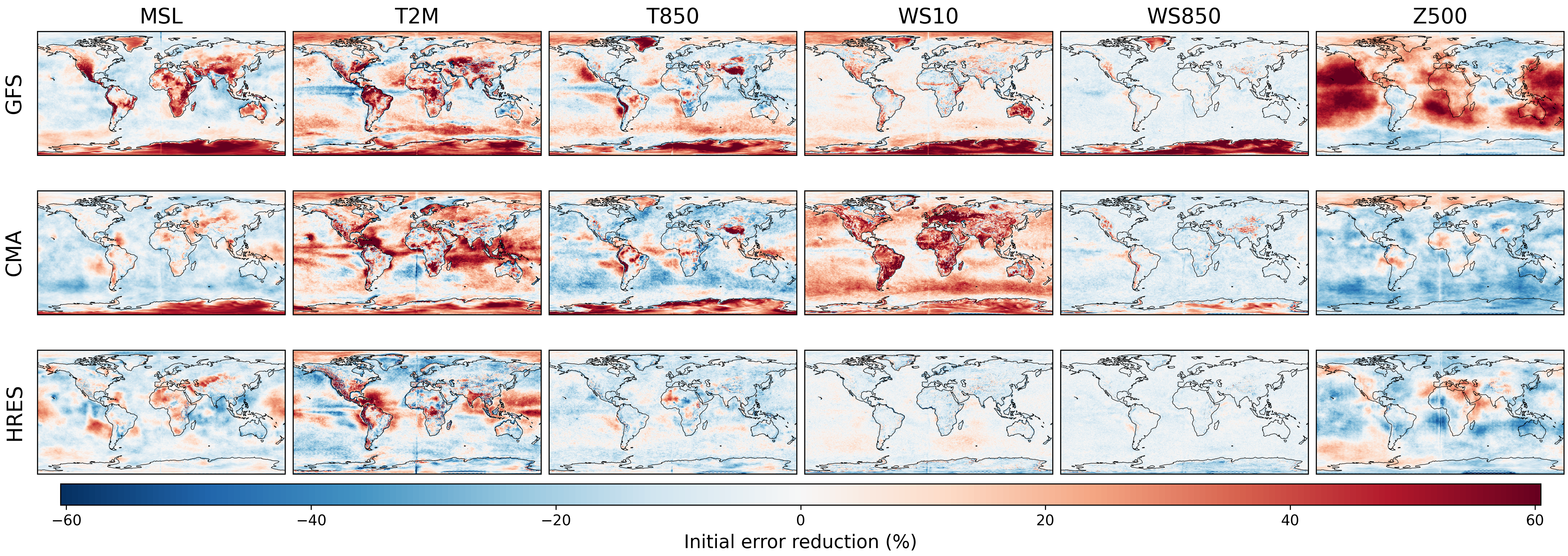}
\caption{Reduction of the initial-field error against ERA5 obtained by the WeatherBridge-aligned initialization relative to the raw-source initialization, averaged over the 00z initializations of the test period. Rows are sources and columns are variables; red marks grid cells where the aligned initial field is closer to ERA5 than the raw-source field, blue marks cells where it is farther. The map compares initial fields rather than forecasts.}
\label{fig:initial_skill_map}
\end{figure}

Figure~\ref{fig:initial_skill_map} shows the spatial distribution of the initial-state error reduction achieved by \model\ relative to the raw-source fields, averaged over all 00Z initializations in the test period. Positive values indicate regions where the aligned field is closer to ERA5, while negative values indicate local degradation. For GFS, substantial error reductions appear across multiple variables, particularly for MSL and T2M. The Z500 map contains both pronounced improvements and degradations, showing that its aggregate improvement is not spatially uniform. CMA exhibits clear regional gains for T2M and WS10, whereas its Z500 error increases over broad areas. The HRES maps remain largely close to zero, apart from localized changes in T2M and Z500, consistent with the relatively small initial discrepancy between HRES and ERA5.

Table~\ref{tab:varwise} summarizes the global alignment results. PSNR improves for all GFS variables, ranging from 2.89\% to 47.56\%, and for five CMA variables, ranging from 1.51\% to 19.80\%. The only notable decrease is CMA Z500 at 1.24\%, while all HRES changes remain within 1.96\%. SSIM improves in 15 of the 18 cases and remains unchanged in the other three. Together with the spatial maps, these results show improved initial-state agreement with ERA5 for GFS and CMA, limited changes for HRES, and occasional localized degradation.

\subsection{Ablation Study}

\begin{wrapfigure}{r}{0.58\textwidth}
\vspace{-1\baselineskip}
\centering
\includegraphics[width=\linewidth]{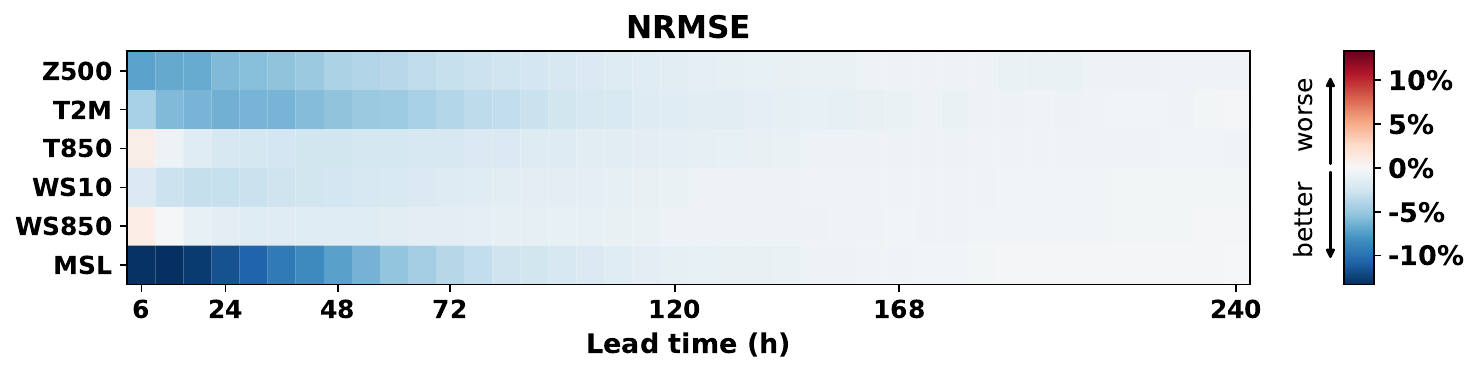}
\vspace{-2\baselineskip}
\caption{NRMSE of GFS-only w/o ORA vs.\ GFS-only; negative (blue) means ORA helps.}
\label{fig:ora_nrmse}
\end{wrapfigure}

We examine the effect of ORA in Figure~\ref{fig:ora_nrmse}.
Compared with reconstruction-only training, ORA provides
the largest RMSE reductions for MSL, followed by clear gains
for T2M and Z500. Gains are smaller for T850 and WS850,
with slight degradation at the first step, and generally
diminish at longer lead times, consistent with the one-step
ORA objective. Source-embedding ablations are provided in
Appendix~\ref{app:ablation}, showing improved 6-hour RMSE
for 13 of 18 source-variable combinations under balanced
training and 17 under source imbalance.

\paragraph{Additional results.} Further ablations, sensitivity tests, and analyses of significance,
scaling, and efficiency appear in Appendix~\ref{app:sec:results}.

\section{Conclusion}
We introduced WeatherBridge, a unified, source-conditioned adapter that maps heterogeneous operational NWP initial fields to ERA5 reference states while leaving the pretrained forecasting model unchanged. WeatherBridge represents the NWP-to-ERA5 mapping as a structured residual correction learned by a shared Swin-UNet, and injects a learnable source embedding at multiple spatial scales so that one adapter captures both shared atmospheric structure and source-specific discrepancies. Its training objective couples latitude-weighted ERA5 reconstruction with One-step Rollout Alignment, which propagates the forecast error of the frozen backbone back into the adapter, so that the adapted states are compatible with the dynamics learned by the backbone rather than merely close to ERA5, so the improvement is not confined to the single training step but holds over the whole 6 h to 240 h rollout.
However, ORA loss that applied to our adapter is one-step: the adapted state is optimized through a single forecast step, whereas forecast skill is ultimately judged over a long auto-regressive rollout. In future work, we plan to extend the alignment object to a multi-step rollout so that adapter is optimized against the error accumulation through long-lead forecasts.

\subsection*{AI use statement}
We used generative AI tools for language editing, LaTeX assistance, table formatting, and literature discovery for related work. We did not use generative AI tools to generate or alter data, run experiments, replace author decisions about the method, or make unsupported scientific claims. All AI-assisted text and code were reviewed by the authors. The authors designed the method and experiments, executed and tested the code, checked all
numerical results against the recorded outputs, and manually verified the relevant literature. We take responsibility for the final content of this work, including any text, claims, code, or other artifacts produced with the aid of generative AI.

\subsection*{Ethics statement}
This work uses established meteorological reanalysis and numerical weather prediction datasets, including ERA5 and operational forecast sources, together with released pretrained weather forecasting models. It involves no human-subject data collection, personally identifiable information, or newly collected sensitive data. The datasets describe atmospheric and environmental conditions rather than individual-level information. Nevertheless, differences in spatial coverage, data quality, physical parameterizations, and source-model characteristics may introduce systematic biases into the learned alignment and downstream forecasts. We therefore recommend validating the proposed method on the target region, variables, and operational setting before deployment, particularly where forecasts may inform safety-critical decisions. We report the data sources, model configurations, and evaluation protocols used in this work to support reproducibility and responsible use.

\subsection*{Reproducibility statement}

The paper and appendix provide the complete method specification, the dataset and split
descriptions, the preprocessing and normalization choices, the model and optimization settings, and the evaluation protocol, including the initialization paths, the forecast horizon, the verification variables and the metric definitions. All reported results follow a single deterministic evaluation protocol applied to the trained adapter weights, and the significance of the main improvements is established by a paired resampling test over the evaluation initializations. The pretrained forecasting models are used exactly as released and are never updated, both in the main experiments and in the cross-backbone transfer test, so no reported result requires retraining a forecasting model. Upon acceptance, we will release the complete source code, configuration files, preprocessing and evaluation scripts, the trained adapter for every configuration reported, and the run metadata needed to reproduce the experiments; the release will cover the multi-source adapter as well as the ablation, sensitivity and cross-backbone transfer settings.

\bibliography{iclr2027_conference}
\bibliographystyle{iclr2027_conference}

\clearpage
\appendix

\section{Implementation and evaluation details}

\subsection{Training Variables List}

Table~\ref{tab:trainingvars} summarizes the 70 channels used by our model: 65 upper-air channels comprising five variables (\texttt{z}, \texttt{t}, \texttt{u}, \texttt{v}, \texttt{r}) at 13 pressure levels (50, 100, 150, 200, 250, 300, 400, 500, 600, 700, 850, 925, and 1000\,hPa), and five surface variables (\texttt{t2m}, \texttt{u10m}, \texttt{v10m}, \texttt{msl}, \texttt{tp}). All source and ERA5 fields are normalized using the same ERA5 per-channel means and standard deviations. The wind speeds \texttt{ws10} and \texttt{ws850} are computed as the magnitudes of the horizontal wind vectors from the corresponding U and V components at 10\,m and 850\,hPa, respectively.

\begin{table}[!t]
\caption{A summary of the 70 channels used by WeatherBridge. The Units column gives the physical unit of each channel.}
\label{tab:trainingvars}
\centering
\footnotesize
\begin{tabular}{lllclc}
\toprule
Type & Full name & Abbreviation & Role & Units & Channels \\
\midrule
\multirow{5}{*}{Upper-air variables} & Geopotential & Z & Input and output & $\mathrm{m^2\,s^{-2}}$ & 13 \\
 & Temperature & T & Input and output & $\mathrm{K}$ & 13 \\
 & u component of wind & U & Input and output & $\mathrm{m\,s^{-1}}$ & 13 \\
 & v component of wind & V & Input and output & $\mathrm{m\,s^{-1}}$ & 13 \\
 & Relative humidity & R & Input and output & \% & 13 \\
\midrule
\multirow{5}{*}{Surface variables} & 2-m temperature & T2M & Input and output & $\mathrm{K}$ & 1 \\
 & 10-m u wind component & U10M & Input and output & $\mathrm{m\,s^{-1}}$ & 1 \\
 & 10-m v wind component & V10M & Input and output & $\mathrm{m\,s^{-1}}$ & 1 \\
 & Mean sea-level pressure & MSL & Input and output & $\mathrm{Pa}$ & 1 \\
 & Total precipitation & TP & Input and output & $\mathrm{m}$ & 1 \\
\midrule
\multicolumn{5}{@{}l}{Total} & 70 \\
\bottomrule
\end{tabular}
\end{table}

\subsection{Channel List and Normalization}
\label{app:channels}

Table~\ref{tab:ref_rmse} lists the nine guided channels of the ORA loss and the per-channel reference error $\mathrm{RMSE}_{\mathrm{ref}}^g$ used to normalize them; the normalization is defined in Section 3.3. 

\begin{table}[!t]
\caption{Per-channel reference error $\mathrm{RMSE}_{\mathrm{ref}}^g$ used to normalize the ORA loss: the RMSE of the frozen backbone initialized from ERA5 at the first inference step.}
\label{tab:ref_rmse}
\centering
\begin{tabular}{lc}
\toprule
Channel & Per-channel RMSE \\
\midrule
Z500 & 21.8412 \\
Z850 & 22.2562 \\
U200 & 1.0605 \\
T2M & 0.6205 \\
U10M & 0.4522 \\
V10M & 0.4659 \\
MSL & 31.0368 \\
TP & 0.6198 \\
T300 & 0.2347 \\
\bottomrule
\end{tabular}
\end{table}

\subsection{Evaluation Metrics}
\label{app:metrics}
\begin{equation}\mathrm{RMSE}_c(\tau) = \sqrt{ \frac{\sum_i \cos(\phi_i) \left( f_{c,i}(\tau) - o_{c,i}(\tau) \right)^2}{\sum_i \cos(\phi_i)} }\label{eq:rmse}\end{equation}
\begin{equation}\mathrm{ACC}_c(\tau) = \frac{\sum_i \cos(\phi_i) f'_{c,i}(\tau) \, o'_{c,i}(\tau)}{\sqrt{ \sum_i \cos(\phi_i) f'_{c,i}(\tau)^2 \; \cdot \; \sum_i \cos(\phi_i) o'_{c,i}(\tau)^2 }}\label{eq:acc}\end{equation}
RMSE and ACC are computed per variable, per lead time and per source; $\cos(\phi)$ provides the latitude weighting.
Here $f$ and $o$ are the forecast and ERA5 truth fields, $f'$ and $o'$ are anomaly fields relative to the climatology, $\tau$ is the lead time, and $\cos(\phi)$ provides area weighting.

Lower RMSE indicates smaller errors; higher ACC indicates better agreement in anomaly patterns. Following FuXi-RTM~\citep{huang2025fuxirtm}, we express the RMSE difference of \model\ relative to the baseline as
{\small
\begin{equation}
    \operatorname{NRMSE}^s_c(n)=
    \frac{\operatorname{RMSE}^s_c(n)-\operatorname{RMSE}^s_{c,\mathrm{base}}(n)}
    {\operatorname{RMSE}^s_{c,\mathrm{base}}(n)}\times100\%.
    \label{eq:eval-nrmse}
\end{equation}
}
Thus, negative NRMSE indicates improvement over the baseline.

PSNR and SSIM assess the aligned initial field $\hat{x}_t^E$ against the ERA5 field $x_t^E$ at the same valid time, per variable and per source, and are computed over the whole field with a single window, without latitude weighting:
\begin{equation}\mathrm{PSNR}_c = 20 \log_{10} \left( \frac{L_c}{\sqrt{\frac{1}{N} \sum_i \left( \hat{x}_{c,i}^{E} - x_{c,i}^{E} \right)^2}} \right)\label{eq:psnr}\end{equation}
\begin{equation}\mathrm{SSIM}_c = \frac{\left( 2 \mu_{\hat{x}} \mu_{x} + C_1 \right) \left( 2 \sigma_{\hat{x} x} + C_2 \right)}{\left( \mu_{\hat{x}}^2 + \mu_{x}^2 + C_1 \right) \left( \sigma_{\hat{x}}^2 + \sigma_{x}^2 + C_2 \right)}\label{eq:ssim}\end{equation}
where $\mu$ and $\sigma$ are the spatial mean and standard deviation of the aligned field $\hat{x}$ and of the ERA5 field $x$, $\sigma_{\hat{x}x}$ is their spatial covariance, and $N$ is the number of grid points.
$L_c$ is the data range of variable $c$, defined as the difference between the 1st and 99th percentiles of the ERA5 field, and it also sets the SSIM constants $C_1 = (0.01\,L_c)^2$ and $C_2 = (0.03\,L_c)^2$.

\subsection{ERA5 Reference Initialization}
\label{app:era5ref}
Table~\ref{tab:era5_ref} reports the reference path, in which the frozen backbone is driven by ERA5 fields, per evaluation variable. It is the best initialization available in this setup and therefore serves as the reference against which raw-source and WeatherBridge-aligned initialization are compared.

\begin{table}[t]
\caption{ERA5 reference initialization: the frozen backbone driven by ERA5 fields, per evaluation variable, at 6 h and 240 h.}
\label{tab:era5_ref}
\centering
\small
\setlength{\tabcolsep}{4pt}
\begin{tabular}{lcccc}
\toprule
Variable & RMSE (6 h) & ACC (6 h) & RMSE (240 h) & ACC (240 h) \\
\midrule
Z500 & 21.71 & 0.9996 & 745.90 & 0.5420 \\
T2M & 0.60 & 0.9825 & 2.67 & 0.5937 \\
T850 & 0.38 & 0.9943 & 3.49 & 0.4757 \\
WS10 & 0.46 & 0.9841 & 2.97 & 0.2774 \\
WS850 & 0.66 & 0.9892 & 5.10 & 0.2853 \\
MSL & 32.02 & 0.9990 & 686.49 & 0.4975 \\
\bottomrule
\end{tabular}
\end{table}

\section{Extended Results}
\label{app:sec:results}
\subsection{Statistical Significance of the Main Results}
\label{app:significance}

We also performed significance tests on the improvements reported in Section~\ref{subsec:weather_forecasting}. For each of the 326 daily 00z initializations shared between the raw-source and \model-aligned runs, we compute the mean over the 40 six-hourly lead steps, form the per-initialization relative improvement, and bootstrap its mean (10{,}000 resamples) to obtain a 95\% percentile confidence interval (Table~\ref{tab:significance}). 
The intervals exclude zero for 15 of the 18 source-variable pairs on RMSE and for 13 of the 18 pairs on ACC: all six GFS and all six CMA variables are significant, whereas among the HRES variables only T2M, T850 and WS10 reach significance on RMSE, consistent with HRES already starting close to the ERA5 reference and therefore having little headroom. The three RMSE pairs whose interval contains zero (Z500, WS850 and MSL on HRES) differ by less than 0.2\% in the mean.

\begin{table}[t]
\caption{Paired bootstrap confidence intervals for the mean rollout RMSE improvement over the raw-source initialization. Each initialization's mean over the 40 six-hourly lead steps (6\,h to 240\,h) is formed first and the improvement is then averaged over the 326 daily $00$z initializations; the last column marks whether the 95\% percentile interval (10{,}000 resamples) excludes zero. The pooled mean in Table~\ref{tab:varwise} weights every lead step equally instead, so the two averages differ slightly. ACC behaves the same way: 13 of the 18 pairs have an interval excluding zero.}
\label{tab:significance}
\centering
\footnotesize
\setlength{\tabcolsep}{4pt}

\begin{tabular}{llccc}
\toprule
Source & Var & Mean RMSE improvement (\%) & 95\% CI & Sig. \\
\midrule
\multirow{6}{*}{GFS} & Z500 & +46.11 & $[+45.50, +46.75]$ & yes \\
 & T2M & +45.70 & $[+45.14, +46.24]$ & yes \\
 & T850 & +41.21 & $[+40.76, +41.65]$ & yes \\
 & WS10 & +32.60 & $[+32.30, +32.91]$ & yes \\
 & WS850 & +30.54 & $[+30.20, +30.88]$ & yes \\
 & MSL & +41.75 & $[+41.12, +42.40]$ & yes \\
\multirow{6}{*}{HRES} & Z500 & -0.17 & $[-0.64, +0.28]$ & no \\
 & T2M & +0.72 & $[+0.50, +0.95]$ & yes \\
 & T850 & +0.57 & $[+0.33, +0.80]$ & yes \\
 & WS10 & +0.17 & $[+0.00, +0.33]$ & yes \\
 & WS850 & -0.00 & $[-0.18, +0.17]$ & no \\
 & MSL & +0.18 & $[-0.26, +0.61]$ & no \\
\multirow{6}{*}{CMA} & Z500 & +9.17 & $[+8.15, +10.18]$ & yes \\
 & T2M & +25.87 & $[+25.11, +26.61]$ & yes \\
 & T850 & +9.97 & $[+9.39, +10.54]$ & yes \\
 & WS10 & +5.72 & $[+5.37, +6.06]$ & yes \\
 & WS850 & +4.14 & $[+3.80, +4.48]$ & yes \\
 & MSL & +19.35 & $[+18.29, +20.41]$ & yes \\
\bottomrule
\end{tabular}
\end{table}

\subsection{Ablation Study}
\label{app:ablation}

In this experiment setting the ORA weight to zero is equivalent to w/o ORA loss, and it degrades the rollout: with both models trained on GFS only, the 6\,h RMSE increases for 4 of the 6 evaluation variables, by 2.0\% to 15.1\%, and changes by less than 1\% for the remaining two (T850 and WS850, where it slightly decreases), and the measured ORA loss rises from 1.8491 to 1.9702. Removing the source embedding also degrades the rollout: the 6\,h RMSE increases for 13 of the 18 source-variable pairs, by up to 8.0\%, and its contribution is largest under source imbalance, where removing it raises the 6\,h RMSE for 17 of the 18 pairs and by up to 13.9\% (Tables~\ref{apptab:ablation} and \ref{tab:stress}).

\begin{table}[t]
\caption{Ablation study. Each variant is compared with the full model trained on the same sources: the ORA-weight ablation is trained on GFS only, and the source-embedding ablation is trained on GFS, HRES and CMA.}
\label{apptab:ablation}
\centering
\footnotesize
\setlength{\tabcolsep}{3pt}
\begin{tabular}{llccccccccc}
\toprule
\multirow{2}{*}{Source} & \multirow{2}{*}{Setting} & \multirow{2}{*}{SSIM} & \multirow{2}{*}{PSNR} & \multicolumn{6}{c}{RMSE (6\,h) $\downarrow$} & \multirow{2}{*}{ORA loss} \\
\cmidrule(lr){5-10}
 & & & & Z500 & T2M & T850 & WS10 & WS850 & MSL & \\
\midrule
\multicolumn{11}{@{}l}{ORA weight ablation (GFS only)} \\
\multirow{2}{*}{GFS} & $\lambda_f = 8\times10^{-3}$ (default) & 0.9839 & 32.70 & 42.20 & 1.04 & 0.91 & 0.88 & 1.51 & 70.98 & 1.8491 \\
 & $\lambda_f = 0$ & 0.9844 & 32.57 & 45.43 & 1.08 & 0.90 & 0.90 & 1.49 & 81.67 & 1.9702 \\
\midrule
\multicolumn{11}{@{}l}{Source-embedding ablation (GFS, HRES and CMA)} \\
\multirow{2}{*}{GFS} & Full WeatherBridge & 0.9851 & 32.99 & 42.07 & 1.01 & 0.88 & 0.88 & 1.49 & 69.20 & 1.7574 \\
 & w/o source embedding & 0.9851 & 32.97 & 40.20 & 1.03 & 0.87 & 0.91 & 1.46 & 74.73 & 1.7718 \\
\multirow{2}{*}{HRES} & Full WeatherBridge & 0.9904 & 35.81 & 29.39 & 0.81 & 0.70 & 0.72 & 1.19 & 46.80 & 1.7574 \\
 & w/o source embedding & 0.9903 & 35.76 & 29.33 & 0.82 & 0.71 & 0.72 & 1.20 & 47.58 & 1.7718 \\
\multirow{2}{*}{CMA} & Full WeatherBridge & 0.9827 & 32.18 & 53.85 & 1.10 & 1.03 & 0.93 & 1.63 & 77.26 & 1.7574 \\
 & w/o source embedding & 0.9826 & 32.12 & 53.49 & 1.11 & 1.04 & 0.93 & 1.63 & 77.78 & 1.7718 \\
\bottomrule
\end{tabular}
\end{table}

Degradation under imbalance scales with the training share: for the scarce CMA source, PSNR/SSIM drop from 32.18/0.9827 (balanced) to 31.73/0.9808 (imbalanced) and further to 31.14/0.9782 when the source embedding is removed (imb\_wo\_src), while the dominant HRES source is least affected (PSNR 35.81, 35.71, 35.47). Removing the embedding under imbalance degrades 17 of the 18 source-variable pairs at 6\,h, by up to 13.9\%, so the source embedding contributes most when the sources are imbalanced (Table~\ref{tab:stress}). By 240\,h these differences have largely disappeared: the three training settings differ by at most 1.0\% on any source-variable pair, and the ordering seen at 6\,h no longer holds, so the imbalance penalty is an early-rollout effect that error growth masks at long lead times. Adding the source embedding leaves 5 of the 18 source-variable pairs slightly worse, but it lowers the overall ORA loss, which averages the per-channel RMSE normalized by the ERA5 reference error over the guided channels, from 1.7718 to 1.7574. The embedding therefore favours the aggregate multi-variable error over any single variable, so individual variables may move in either direction while the overall error still decreases.

\begin{table}[t]
\caption{Source-imbalance stress test: balanced full-source training, imbalanced training (HRES 100\%, GFS 50\%, CMA 10\%), and imbalanced training without the source embedding. PSNR/SSIM describe the WeatherBridge-aligned initial field against ERA5, so they do not depend on lead time and are listed once, in the upper block; the upper block then gives the RMSE of the six evaluation variables at the first step (6\,h) and the lower block the same RMSE at 240\,h, where forecast error has largely saturated.}
\label{tab:stress}
\centering
\footnotesize
\setlength{\tabcolsep}{3pt}
\begin{tabular}{llcccccccc}
\toprule
\multirow{2}{*}{Source} & \multirow{2}{*}{Training mix} & \multirow{2}{*}{PSNR} & \multirow{2}{*}{SSIM} & \multicolumn{6}{c}{RMSE (6\,h) $\downarrow$} \\
\cmidrule(lr){5-10}
 & & & & Z500 & T2M & T850 & WS10 & WS850 & MSL \\
\midrule
\multirow{3}{*}{GFS} & Balanced & 32.99 & 0.9851 & 42.07 & 1.01 & 0.88 & 0.88 & 1.49 & 69.20 \\
 & Imbalanced & 32.62 & 0.9840 & 42.04 & 1.03 & 0.91 & 0.89 & 1.50 & 71.13 \\
 & Imbalanced, w/o src emb & 31.94 & 0.9823 & 43.55 & 1.12 & 0.97 & 0.91 & 1.53 & 79.29 \\
\multirow{3}{*}{HRES} & Balanced & 35.81 & 0.9904 & 29.39 & 0.81 & 0.70 & 0.72 & 1.19 & 46.80 \\
 & Imbalanced & 35.71 & 0.9900 & 28.89 & 0.82 & 0.71 & 0.72 & 1.20 & 46.97 \\
 & Imbalanced, w/o src emb & 35.47 & 0.9897 & 29.90 & 0.86 & 0.76 & 0.74 & 1.22 & 49.96 \\
\multirow{3}{*}{CMA} & Balanced & 32.18 & 0.9827 & 53.85 & 1.10 & 1.03 & 0.93 & 1.63 & 77.26 \\
 & Imbalanced & 31.73 & 0.9808 & 53.24 & 1.17 & 1.06 & 0.95 & 1.67 & 79.17 \\
 & Imbalanced, w/o src emb & 31.14 & 0.9782 & 51.19 & 1.33 & 1.15 & 0.99 & 1.70 & 84.00 \\
\midrule
\multicolumn{10}{@{}l}{RMSE (240\,h) $\downarrow$} \\
\midrule
\multirow{3}{*}{GFS} & Balanced & & & 790.22 & 2.78 & 3.65 & 3.05 & 5.23 & 727.50 \\
 & Imbalanced & & & 788.96 & 2.78 & 3.64 & 3.05 & 5.24 & 727.58 \\
 & Imbalanced, w/o src emb & & & 784.68 & 2.80 & 3.63 & 3.04 & 5.21 & 722.84 \\
\multirow{3}{*}{HRES} & Balanced & & & 747.00 & 2.68 & 3.48 & 2.98 & 5.10 & 691.29 \\
 & Imbalanced & & & 746.09 & 2.68 & 3.47 & 2.98 & 5.11 & 691.46 \\
 & Imbalanced, w/o src emb & & & 747.26 & 2.68 & 3.48 & 2.98 & 5.11 & 692.05 \\
\multirow{3}{*}{CMA} & Balanced & & & 806.73 & 2.83 & 3.71 & 3.07 & 5.28 & 742.12 \\
 & Imbalanced & & & 798.97 & 2.82 & 3.69 & 3.07 & 5.28 & 736.94 \\
 & Imbalanced, w/o src emb & & & 802.00 & 2.84 & 3.70 & 3.08 & 5.29 & 741.08 \\
\bottomrule
\end{tabular}
\end{table}

\subsection{Parameter Sensitivity}
\label{app:param}

\begin{figure}[!t]
\centering
\includegraphics[width=\linewidth]{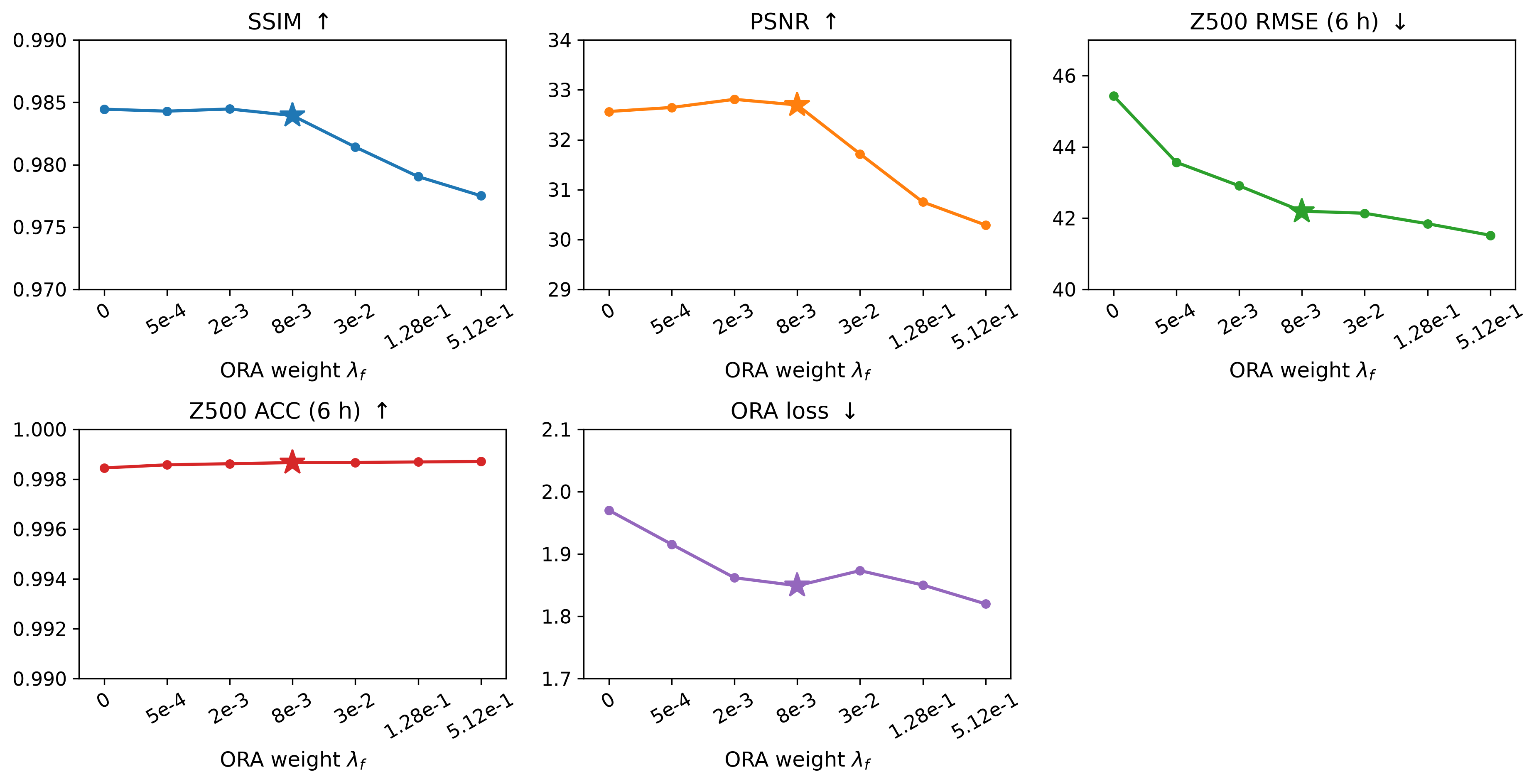}
\caption{Sensitivity to the ORA weight $\lambda_f$ (default $8\times10^{-3}$; source: GFS). Each panel shows one metric against the weight: initial-field SSIM and PSNR, first-step Z500 RMSE and ACC, and the ORA loss; the star marks the default setting.}
\label{fig:sensitivity}
\end{figure}

Increasing the ORA weight from 0 to the default 8e-3 lowers the first-step Z500 RMSE from 45.43 to 42.20 while Z500 ACC stays around 0.999; SSIM/PSNR peak near 2e-3 (0.9845, 32.81) and degrade sharply beyond 1.28e-1 (SSIM 0.9790, PSNR 30.75). The default 8e-3 therefore sits on the fidelity-utility frontier (Figure~\ref{fig:sensitivity}).

\subsection{Multi-Source Training Structure (Source Mixing)}
\label{app:sourcemix}

Different mixtures of sources give different results (Table~\ref{tab:sourcemix}). Evaluated on a given source, training together with a second source usually beats training on that source alone: for GFS, adding HRES cuts the 6 h RMSE of the most improved variable, T850, by 2.2\% and degrades none of the six, for a mean reduction of 1.4\%, while adding CMA instead leaves three of the six variables slightly worse, by at most 1.1\%, and the mean 0.4\% higher. Over all sources the worst pair is still only 2.9\% behind the single-source model. The gain is largest for CMA, the weakest source on its own, which improves on five of its six variables when trained together with HRES. We therefore train the final adapter on all three sources, since no negative interference between sources is observed.

\begin{table}[t]
\caption{Multi-source training structure: single-source models and pair-trained adapters, each evaluated on every source. }
\label{tab:sourcemix}
\centering
\footnotesize
\setlength{\tabcolsep}{3pt}
\begin{tabular}{llcccccccc}
\toprule
\multirow{2}{*}{Eval source} & \multirow{2}{*}{Training setup} & \multirow{2}{*}{PSNR} & \multirow{2}{*}{SSIM} & \multicolumn{6}{c}{RMSE (6\,h) $\downarrow$} \\
\cmidrule(lr){5-10}
 & & & & Z500 & T2M & T850 & WS10 & WS850 & MSL \\
\midrule
\multirow{3}{*}{GFS} & GFS-only & 32.70 & 0.9839 & 42.20 & 1.04 & 0.91 & 0.88 & 1.51 & 70.98 \\
 & GFS+CMA & 32.67 & 0.9842 & 42.51 & 1.04 & 0.91 & 0.89 & 1.51 & 71.32 \\
 & GFS+HRES & 32.82 & 0.9844 & 41.35 & 1.03 & 0.89 & 0.88 & 1.49 & 69.60 \\
\multirow{3}{*}{HRES} & HRES-only & 35.67 & 0.9898 & 29.01 & 0.82 & 0.72 & 0.72 & 1.21 & 47.06 \\
 & GFS+HRES & 35.72 & 0.9901 & 29.28 & 0.82 & 0.72 & 0.72 & 1.20 & 47.17 \\
 & CMA+HRES & 35.81 & 0.9904 & 29.16 & 0.81 & 0.71 & 0.72 & 1.19 & 46.72 \\
\multirow{3}{*}{CMA} & CMA-only & 31.87 & 0.9812 & 51.99 & 1.17 & 1.06 & 0.95 & 1.66 & 77.62 \\
 & GFS+CMA & 31.95 & 0.9815 & 53.49 & 1.14 & 1.05 & 0.94 & 1.66 & 78.03 \\
 & CMA+HRES & 32.17 & 0.9828 & 53.06 & 1.10 & 1.03 & 0.93 & 1.62 & 77.07 \\
\bottomrule
\end{tabular}
\end{table}

\subsection{Data Scale}
\label{app:datascale}

Rollout and initial-field metrics improve monotonically with more paired data: SSIM 0.9795 to 0.9839, PSNR 31.12 to 32.70, and the 6\,h RMSE of every evaluation variable falls from 25\% to 100\% of the samples, for example from 44.45 to 42.20 for Z500 and from 83.97 to 70.98 for MSL, showing that the source-to-ERA5 structured bias is learnable and that WeatherBridge scales with data (Table~\ref{tab:datascale}).

\begin{table}[t]
\caption{Data-scale study (default source GFS): training on 25\%, 50\% and 100\% of the paired samples.}
\label{tab:datascale}
\centering
\small
\setlength{\tabcolsep}{4pt}
\begin{tabular}{lccccccccc}
\toprule
\multirow{2}{*}{Training samples} & \multirow{2}{*}{SSIM} & \multirow{2}{*}{PSNR} & \multicolumn{6}{c}{RMSE (6\,h) $\downarrow$} & \multirow{2}{*}{ORA loss} \\
\cmidrule(lr){4-9}
 & & & Z500 & T2M & T850 & WS10 & WS850 & MSL & \\
\midrule
25\% & 0.9795 & 31.12 & 44.45 & 1.20 & 1.03 & 0.92 & 1.54 & 83.97 & 2.0340 \\
50\% & 0.9803 & 31.45 & 44.46 & 1.17 & 1.00 & 0.92 & 1.54 & 83.57 & 2.0160 \\
100\% & 0.9839 & 32.70 & 42.20 & 1.04 & 0.91 & 0.88 & 1.51 & 70.98 & 1.8491 \\
\bottomrule
\end{tabular}
\end{table}

\subsection{Efficiency and Training Cost}
\label{app:efficiency}

The smallest configuration is the best: Lite (138.4M) attains the lowest 6\,h RMSE on all six evaluation variables and the lowest ORA loss (1.8491), while Mid (222.5M) and Deep (285.4M) are slightly worse (for example from 1.04 to 1.08 and 1.11 for T2M, and from 70.98 to 71.89 and 74.35 for MSL). Capacity scaling therefore saturates, which supports using the Lite model as the main configuration and treating WeatherBridge as a lightweight front-end adapter (Table~\ref{tab:efficiency}). The Lite model trains in about 88 h on 4 A100 GPUs (about 350 GPU-hours), and the test-time FuXi rollout over the three sources (326 initializations each) takes about 46 h in total.

\begin{table}[t]
\caption{Model-scale study (default source: GFS). }
\label{tab:efficiency}
\centering
\small
\setlength{\tabcolsep}{5pt}
\begin{tabular}{lccccccccc}
\toprule
\multirow{2}{*}{Model} & \multirow{2}{*}{SSIM} & \multirow{2}{*}{PSNR} & \multicolumn{6}{c}{RMSE (6\,h) $\downarrow$} & \multirow{2}{*}{ORA loss} \\
\cmidrule(lr){4-9}
 & & & Z500 & T2M & T850 & WS10 & WS850 & MSL & \\
\midrule
Lite & 0.9839 & 32.70 & 42.20 & 1.04 & 0.91 & 0.88 & 1.51 & 70.98 & 1.8491 \\
Mid & 0.9836 & 32.50 & 42.65 & 1.08 & 0.92 & 0.89 & 1.52 & 71.89 & 1.8793 \\
Deep & 0.9831 & 32.33 & 43.25 & 1.11 & 0.94 & 0.90 & 1.52 & 74.35 & 1.8986 \\
\bottomrule
\end{tabular}
\end{table}

\subsection{Transfer to a Second Backbone (FengWu)}
\label{app:subsec:fengwu}
WeatherBridge is trained jointly with a frozen FuXi backbone. In this experiment the FengWu backbone is used as released: only its initial fields are replaced by the WeatherBridge-aligned states, so the adapter is applied unchanged and nothing is retrained for FengWu. Table~\ref{tab:fengwu_transfer} reports the resulting rollout on the GFS source, which shows that the adapted states are usable by a different ERA5-trained backbone.

\begin{table}[!t]
\caption{Naive and WeatherBridge-aligned initialization on FengWu for the GFS source: RMSE and ACC at 6\,h and 240\,h per variable, with the raw-source value followed by the WeatherBridge-aligned value and by $\Delta$, the relative improvement of the \model-aligned result over the raw-source result in percent, computed before rounding; ``$\uparrow$'' and ``$\downarrow$'' indicate that higher and lower values are better, respectively.}
\label{apptab:fengwu_transfer}
\centering
\scriptsize
\setlength{\tabcolsep}{1.5pt}
\resizebox{\textwidth}{!}{%
\begin{tabular}{
@{}l|cc>{\columncolor{gray!15}}c|cc>{\columncolor{gray!15}}c|cc>{\columncolor{gray!15}}c|cc>{\columncolor{gray!15}}c@{}
}
\toprule
\multirow{2}{*}{Var} & \multicolumn{3}{c|}{RMSE 6\,h $\downarrow$} & \multicolumn{3}{c|}{ACC 6\,h $\uparrow$} & \multicolumn{3}{c|}{RMSE 240\,h $\downarrow$} & \multicolumn{3}{c}{ACC 240\,h $\uparrow$} \\
\cmidrule(lr){2-4}\cmidrule(lr){5-7}\cmidrule(lr){8-10}\cmidrule(lr){11-13}
 & raw & WeatherBridge & $\Delta$ & raw & WeatherBridge & $\Delta$ & raw & WeatherBridge & $\Delta$ & raw & WeatherBridge & $\Delta$ \\
\midrule
Z500 & 226.94 & 56.41 & 75.14\% & 0.9621 & 0.9976 & 3.70\% & 1023.32 & 952.65 & 6.91\% & 0.0732 & 0.1748 & 138.68\% \\
T2M & 2.28 & 1.22 & 46.24\% & 0.7485 & 0.9268 & 23.82\% & 3.93 & 3.23 & 17.77\% & 0.3584 & 0.4365 & 21.78\% \\
T850 & 6.14 & 0.93 & 84.83\% & 0.2846 & 0.9645 & 238.90\% & 4.81 & 4.08 & 15.25\% & 0.0783 & 0.1717 & 119.27\% \\
WS10 & 1.65 & 0.90 & 45.49\% & 0.7880 & 0.9381 & 19.05\% & 3.88 & 3.86 & 0.61\% & 0.0479 & 0.0907 & 89.50\% \\
WS850 & 2.77 & 1.56 & 43.57\% & 0.7948 & 0.9383 & 18.06\% & 6.29 & 5.91 & 6.08\% & 0.0322 & 0.0873 & 171.17\% \\
MSL & 297.15 & 80.75 & 72.82\% & 0.9112 & 0.9938 & 9.07\% & 867.90 & 727.29 & 16.20\% & 0.0952 & 0.2912 & 205.81\% \\
\bottomrule
\end{tabular}
}
\end{table}

\end{document}